\documentclass[letterpaper,journal,twoside]{IEEEtran}
\usepackage{amsmath,amsfonts}
\usepackage{algorithmic}
\usepackage{algorithm}
\usepackage{array}
\usepackage[caption=false,font=normalsize,labelfont=sf,textfont=sf]{subfig}
\usepackage{textcomp}
\usepackage{stfloats}
\usepackage{url}
\usepackage{verbatim}
\usepackage{graphicx}
\usepackage{xcolor}
\usepackage{cite}
\usepackage{booktabs}
\usepackage{multirow}
\usepackage[table]{xcolor}
\usepackage{colortbl}
\usepackage{array}
\usepackage{etoolbox}
\def\tablename{\textbf{Table}}
\makeatletter
\patchcmd{\@makecaption}{\scshape}{}{}{}
\makeatletter
\patchcmd{\@makecaption}{\\}{.\ }{}{}
\makeatother
\usepackage{threeparttable}
\usepackage{xcolor}
\usepackage{dashrule}
\usepackage{tikz}

\usepackage{array}
\newcolumntype{C}[1]{>{\centering\arraybackslash}p{#1}}

\definecolor{Gray}{gray}{0.9}
\definecolor{LightOrange}{RGB}{255, 242, 224}
\begin{document}

\title{\textbf{PerlAD: Towards Enhanced Closed-loop End-to-end Autonomous Driving with Pseudo-simulation-based Reinforcement Learning}}

\author{
${\text{Yinfeng Gao}}^{1,2,3*\dag}$, 
$\text{Qichao Zhang}^{3*}$, 
$\text{Deqing Liu}^{3}$,  
$\text{Zhongpu Xia}^{3}$,  
$\text{Guang Li}^{2}$,  
$\text{Kun Ma}^{2}$,  
$\text{Guang Chen}^{2}$, 
$\text{Hangjun Ye}^{2}$,
$\text{Long Chen}^{2}$,
$\text{Da-Wei Ding}^{1\ddag}$, $~\IEEEmembership{Senior Member, IEEE}$
and $\text{Dongbin Zhao}^{3}$, $~\IEEEmembership{Fellow, IEEE}$

\thanks{Manuscript received: December, 2, 2025; Revised: February, 4, 2026; Accepted: March, 7, 2026.}
\thanks{This paper was recommended for publication by Editor Olivier Stasse upon evaluation of the Associate Editor and Reviewers’ comments. 
This work was supported by the National Natural Science Foundation of China under Grant No. 62273035, 
in part by the Beijing Natural Science Foundation-Xiaomi Innovation Joint Fund under Grant No. L253007, 
and by the Beijing Natural Science Foundation under Grant Nos. 4262056, 4242052, and 4252045.}
\thanks{$^{1}$ Yinfeng Gao and Da-Wei Ding are with the School of Automation and Electrical Engineering, University of Science and Technology Beijing, Beijing 100083, China. 
Yinfeng Gao is also with Xiaomi EV and the State Key Laboratory of Multimodal Artificial Intelligence Systems, Institute of Automation, Chinese Academy of Sciences, Beijing 100190, China.
{\tt\footnotesize gaoyinfeng07@gmail.com}}
\thanks{$^{2}$ Guang Li, Kun Ma, Guang Chen, Hangjun Ye, and Long Chen are with Xiaomi EV.
{\tt\footnotesize alwaysunny@gmail.com}}
\thanks{$^{3}$ Qichao Zhang, Deqing Liu, Zhongpu Xia, and Dongbin Zhao are with the State Key Laboratory of Multimodal Artificial Intelligence Systems, Institute of Automation, Chinese Academy of Sciences, Beijing 100190, China, and also with the School of Artificial Intelligence, University of Chinese Academy of Sciences, Beijing 100049, China.{\tt\footnotesize zhangqichao2014@ia.ac.cn}}
\thanks{
$^{*}$ Contribute equally. 
$^{\dag}$ Intern at CASIA \& Xiaomi Embodied Intelligence Team.
$^{\ddag}$ Corresponding author.
{\tt\footnotesize ddaweiauto@163.com}}
\thanks{Digital Object Identifier (DOI): see top of this page.}
\vspace{-8mm}}
\markboth{IEEE ROBOTICS AND AUTOMATION LETTERS. PREPRINT VERSION. ACCEPTED MARCH 2026}%
{Gao \MakeLowercase{\textit{et al.}}: Pseudo-simulation-based RL for E2E AD}


\maketitle
\begin{abstract}
End-to-end autonomous driving policies based on Imitation Learning (IL) often struggle in closed-loop execution due to the misalignment between inadequate open-loop training objectives and real driving requirements. 
While Reinforcement Learning (RL) offers a solution by directly optimizing driving goals via reward signals, the rendering-based training environments introduce the rendering gap and are inefficient due to high computational costs.
To overcome these challenges, we present a novel \underline{P}s\underline{e}udo-simulation-based \underline{RL} method for closed-loop end-to-end autonomous driving, PerlAD.
Based on offline datasets, PerlAD constructs a pseudo-simulation that operates in vector space, enabling efficient, rendering-free trial-and-error training.
To bridge the gap between static datasets and dynamic closed-loop environments, PerlAD introduces a prediction world model that generates reactive agent trajectories conditioned on the ego vehicle's plan.
Furthermore, to facilitate efficient planning, PerlAD utilizes a hierarchical decoupled planner that combines IL for lateral path generation and RL for longitudinal speed optimization.
Comprehensive experimental results demonstrate that PerlAD achieves state-of-the-art performance on the Bench2Drive benchmark, surpassing the previous E2E RL method by 10.29\% in Driving Score without requiring expensive online interactions. 
Additional evaluations on the DOS benchmark further confirm its reliability in handling safety-critical occlusion scenarios.
\end{abstract}

\begin{IEEEkeywords}
Autonomous Vehicle Navigation, Integrated Planning and Learning, Reinforcement Learning.
\end{IEEEkeywords}

\section{Introduction}
\IEEEPARstart{E}{nd-to-end} (E2E) autonomous driving has garnered significant attention in recent years. 
Most mainstream methods rely heavily on Imitation Learning (IL)~\cite{hu2023planning, jiang2023vad, jia2025drivetransformer, gao2019compare}, which is trained on large datasets of expert demonstrations.
\begin{figure}[htp]
\centering
\includegraphics[width=0.45\textwidth]{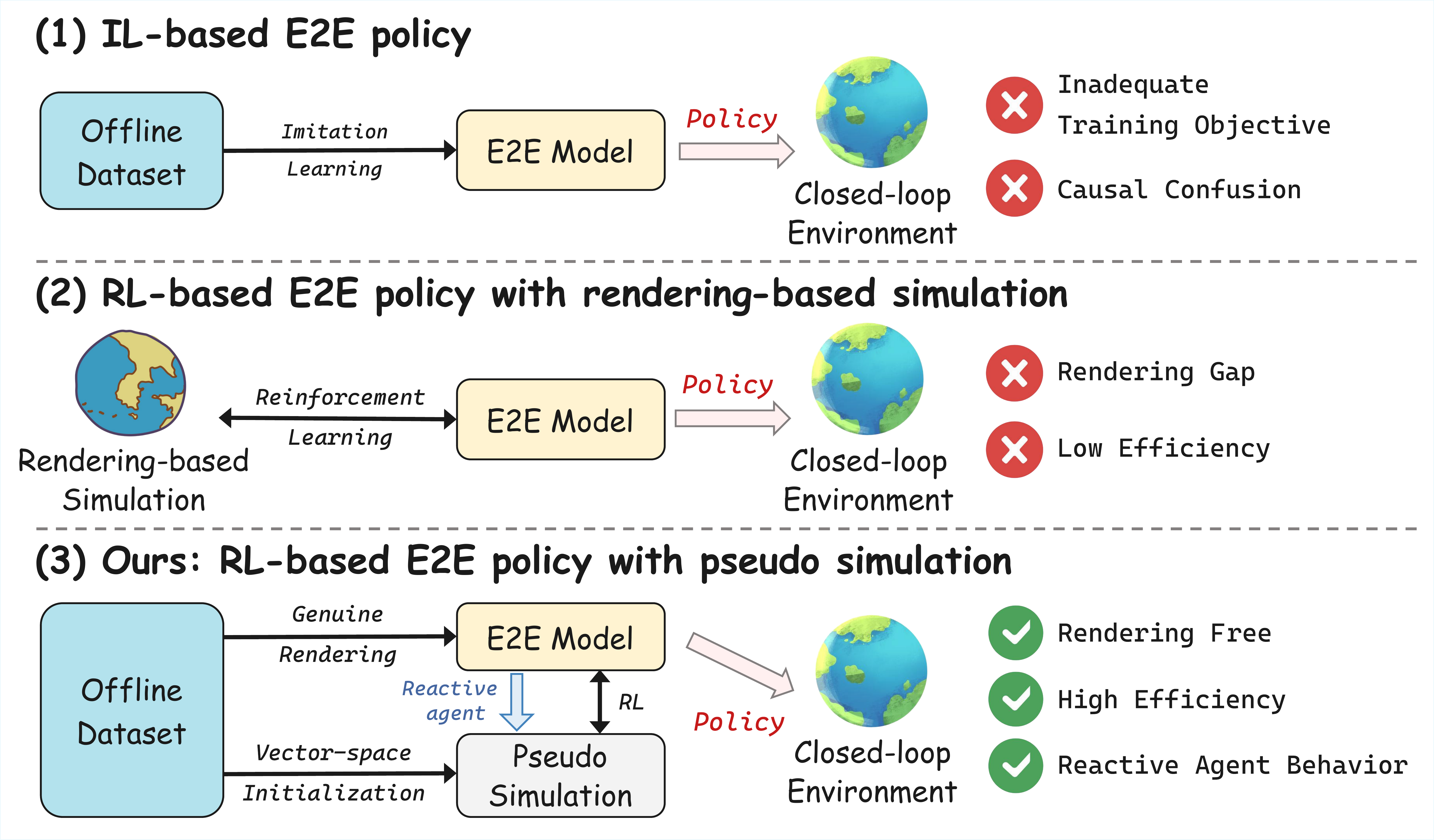}
\vspace{-3mm}
    \caption{Different training paradigms of E2E autonomous driving.}
\label{teasar}
\vspace{-6mm}
\end{figure} 
However, IL methods face fundamental challenges that lead to risky behaviors in closed-loop environments.
The primary issue is the inadequate training objective: IL merely minimizes the geometric deviation between model outputs and demonstrations, which is intrinsically misaligned with real driving requirements, such as safety and efficiency. 
This issue is compounded by causal confusion, which makes IL policies susceptible to learning spurious correlations instead of true causal relationships.
In contrast, Reinforcement Learning (RL) explicitly integrates driving goals through reward modeling~\cite{shi2024task}, enabling trial-and-error exploration to establish causal dependencies. 
Recent works have explored its application in E2E systems.
Nevertheless, existing E2E RL approaches are still hindered by key bottlenecks.
Specifically, some methods propose training E2E RL policies in rendering-based simulation environments, including game engine simulators~\cite{yang2025raw2drive} and sensory reconstruction platforms~\cite{gao2025rad}. 
However, the rendering in the game engine introduces a significant input domain gap between training and deployment, and online interactions with the reconstruction platform demand high computational costs, which makes RL training inefficient.
Alternatively, other RL methods~\cite{li2025learning, yang2025worldrft} attempt to finetune and evaluate E2E policies under an open-loop setting by relying on static, logged agent behaviors. 
Due to these bottlenecks, existing E2E RL methods struggle to achieve competitive performance on public closed-loop benchmarks that require  continuous interaction with the dynamic environment, such as Bench2Drive~\cite{jia2024bench}.

In this work, we present a novel \underline{P}s\underline{e}udo-simulation-based \underline{RL} training method, \textbf{PerlAD}, which is designed for closed-loop E2E autonomous driving and effectively addresses the aforementioned bottlenecks. 
We achieve this by constructing a pseudo-simulation environment that operates in vector space, using real sensor data from offline datasets. 
This approach eliminates complex rendering processes, thereby resolving the input domain gap and boosting training efficiency. 
Furthermore, PerlAD incorporates a Prediction World Model that explicitly predicts surrounding agents' trajectories conditioned on the ego plan. 
This mechanism generates reactive simulations that mimic closed-loop interactions, providing closed-loop consistent reward signals for RL training, distinguishing it from prior driving world models~\cite{gao2024piwm} that use prediction solely for feature extraction in modular systems.
Finally, to facilitate efficient planning, we propose a decoupled planner. 
It combines IL for smooth lateral path generation and RL for interactive longitudinal speed optimization.
The two decoupled actions are then synergistically optimized through an alignment training strategy that balances geometric accuracy and overall driving objectives.
Experiments on the Bench2Drive benchmark~\cite{jia2024bench} demonstrate that PerlAD achieves State-of-The-Art (SoTA) closed-loop performance, surpassing Raw2Drive~\cite{yang2025raw2drive}, which requires expansive online explorations and expert distillation.
Additionally, evaluation on the Driving in Occlusion Simulation (DOS) benchmark~\cite{shao2023reasonnet} verifies PerlAD's efficacy in safely navigating occlusion scenarios.

Our main contributions are summarized as follows:
\begin{enumerate}
\item{We propose \textbf{PerlAD}, an RL training method for closed-loop E2E autonomous driving based on offline datasets, enabling efficient trial-and-error within a pseudo-simulation environment.}
\item{PerlAD integrates a Prediction World Model to generate reactive trajectories of traffic agents, mimicking their interactive behaviors for pseudo-simulation training.}
\item{We introduce a hierarchical, decoupled planning module that leverages RL to optimize complex longitudinal planning tasks, while enhancing lateral path planning quality through an alignment training strategy.}
\item{PerlAD achieves SoTA closed-loop performance on the Bench2Drive benchmark, surpassing the previous E2E RL method that needs expensive online explorations by 10.29\% in Driving Score.
It further demonstrates strong performance on the safety-critical DOS benchmark.}
\end{enumerate}

\section{Related Works}
\subsection{IL-based End-to-end Driving}
E2E autonomous driving aims to directly map raw sensor inputs to ego planning using a single neural network. 
The mainstream approaches primarily leverage the IL paradigm. 
Early works focused on designing efficient network architectures and representations, utilizing unified multi-query frameworks for joint perception, prediction, and planning~\cite{hu2023planning, jiang2023vad, sun2024sparsedrive}.
To address planning uncertainty, some methods explored generating multi-modal trajectories with diffusion-based models~\cite{xing2025mimir, wang2025diffad}.
Furthermore, some work investigated reducing labeled data dependency by developing latent-space world models for self-supervised learning~\cite{zheng2025world4drive}.
Recent approaches leveraged large language models for interpretable reasoning~\cite{shao2024lmdrive, liu2025reasonplan, zheng2026planagent}.
Despite these advances, the above methods rely on IL to minimize the distance between model outputs and expert demonstrations. 
They fail to align the optimization target with high-level driving objectives, such as safety and efficiency, which motivates the shift toward RL paradigms.

\subsection{RL-based End-to-end Driving}
RL has been widely adopted in modular autonomous driving systems with privileged perception, achieving driving performance superior to IL~\cite{zhang2022trajgen, cusumano2025robust, zhang2025carplanner}. 
Recent works have also leveraged RL to optimize E2E models. 
Specifically, some methods train RL policies in rendering-based simulators, including game engines~\cite{yang2025raw2drive} and 3D Gaussian Splatting (3DGS) reconstructions~\cite{gao2025rad}.
However, game-engine rendering introduces domain gaps between simulated and real sensor inputs, while 3DGS-based training suffers from prohibitive computational costs. 
Alternatively, other approaches apply RL for open-loop policy fine-tuning~\cite{li2025learning, yang2025worldrft}, yet these non-reactive settings fail to capture essential dynamic interactions, making them inadequate for assessing closed-loop driving capability. 
To overcome these limitations, we propose PerlAD, which enables efficient RL training through a rendering-free pseudo-simulation environment in vector space, augmented with reactive agent behavior modeling to ensure consistency with dynamic closed-loop scenarios.
Its effectiveness is further validated via comprehensive closed-loop evaluations.

\section{Problem Definition}
E2E driving can be formulated as a Partially Observable Markov Decision Process (POMDP)~\cite{sutton2018reinforcement}, where the policy makes decisions based on partial observations (e.g., sensor inputs) rather than full states (e.g., traffic agents’ motion states). 
A POMDP is defined by the tuple $(\mathcal{X}, \mathcal{S}, \mathcal{A}, \mathcal{T}, \mathcal{O}, \mathcal{R}, \gamma)$, where $\mathcal{X}$, $\mathcal{S}$, and $\mathcal{A}$ are the observation, state, and action spaces.
$\mathcal{T}$, $\mathcal{O}$, and $\mathcal{R}$ are the state transition, observation, and reward functions, and $\gamma$ is the discount factor.
The goal of policy $\pi$ is to maximize the expected cumulative reward.

In E2E autonomous driving, observations $\mathcal{X}$ are derived from sensor inputs. 
In PerlAD, we use surround-view cameras for observation, i.e., $\mathcal{X}=\{x_i\}_{i=1}^{N_{cam}}$, where $x_i$ is the image from camera $i$ and $N_{cam}$ is the number of cameras.
The action space is decoupled into lateral and longitudinal actions~\cite{jaeger2023hidden}, $\mathcal{A}=\{a_{lat}, a_{lon} \}$, where $a_{lat} =\{ w_{lat, i} \}_{i=1}^{N_{lat}}$ is the lateral path planning action, represented by a sequence of equally spaced path waypoints $w_{lat, i}$, $N_{lat}$ is the number of waypoints.
$a_{lon}$ is the longitudinal target speed action, represented as a scalar.
The reward function incorporates terms corresponding to core driving requirements such as safety and efficiency, its detailed formulation is presented in Section \ref{simulation}.


\begin{figure*}[htbp]
    \centering
    \includegraphics[width=0.88\textwidth]{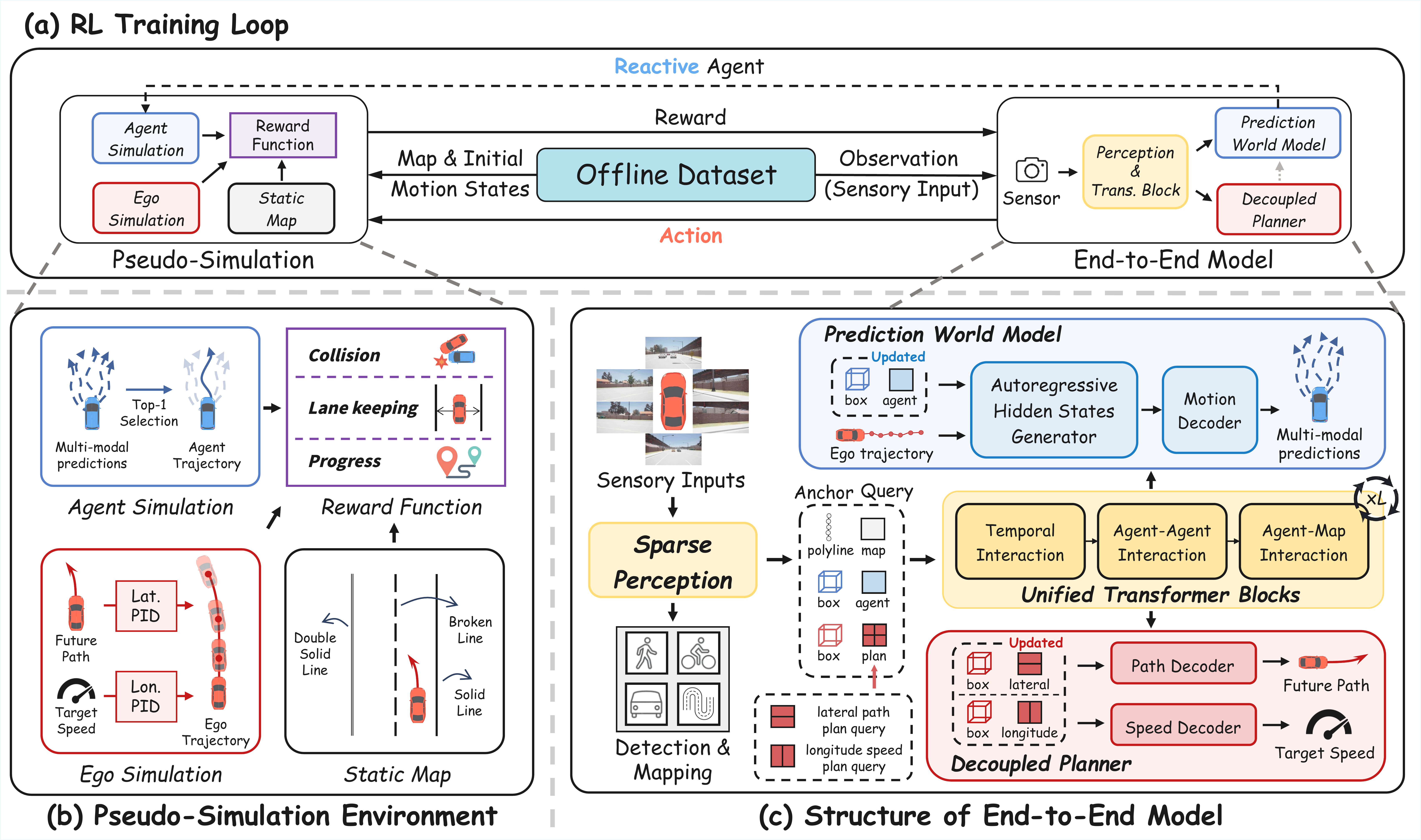}
    \vspace{-3mm}
    \caption{
    The framework of PerlAD.
    (a) The RL training loop.
    An offline dataset initializes motion states for the pseudo-simulation and provides sensor observations for the E2E model.
    The E2E model generates reactive agent predictions and planning actions, which are then provided to the simulation to compute rewards.
    (b) The pseudo-simulation environment.
    It is responsible for simulating future scenarios and calculating rewards.
    (c) The structure of the E2E model. 
    The sparse perception extracts structured representations, which are then processed by unified transformer blocks for feature interaction. 
    This is followed by a decoupled planner that outputs lateral and longitudinal actions, and a prediction world model that generates reactive agent trajectories.
    }
    \label{overview}
    \vspace{-6mm}
\end{figure*}

\section{Method}
PerlAD builds a pseudo-simulation environment entirely from offline datasets for RL training, with the loop illustrated in Fig. \ref{overview}(a). 
Sensor data are extracted as observations and fed into the E2E model, which outputs planning actions and reactive agent trajectories.
These outputs, together with the initial motion state, enter the pseudo-simulation, which simulates the future motion of all traffic participants in vector space and computes rewards. 
The rewards are then fed back to update the RL policy.

\subsection{Pseudo-Simulation Environment}
\label{simulation}
As shown in Fig. \ref{overview}(b), PerlAD builds a rendering-free pseudo-simulation environment that operates entirely in vector space to simulate action outcomes and calculate rewards. 
This environment runs in parallel on the GPU, enabling the policy to learn through efficient trial-and-error.

\subsubsection{\textbf{Ego Simulation}}
In this work, the E2E model's action output is decoupled into lateral and longitudinal planning actions. 
The ego simulation module converts these actions into actual motion trajectories by simulating the ego vehicle's control process. 
Specifically, we assume the ego vehicle follows a bicycle kinematics model.
Its initial motion states are from the dataset, including position, orientation, and velocity. 
Given the lateral path action $a_{lat}$ and longitudinal target speed action $a_{lon}$, the ego simulation module uses two PID controllers to compute accelerations and steering angles, generating the ego's future motion trajectory $P_{ego}^{sim}=\{ w_{ego, t} \}_{t=1}^{T_{sim}}$, where $w_{ego, t}$ denotes the ego's positional waypoint at time $t$, and $T_{sim}$ is the simulation horizon.

\subsubsection{\textbf{Agent Simulation}}
The E2E model generates multi-modal reactive agent predicted trajectories, from which the agent simulation module samples Top-1 trajectory $P_{agent}^{pred}=\{ w_{agent, t} \}_{t=1}^{T_{pred}}$ for simulation, where $w_{agent, t}$ denotes agent's positional waypoint at time $t$, and $T_{pred}$ is the prediction horizon.
Based on this selected trajectory, the agent simulation produces the agent's future motion trajectory $P_{agent}^{sim}=\{ w_{agent, t} \}_{t=1}^{T_{sim}}$.
These two trajectories differ in their time granularity. 
Given the typical sparse temporal resolution of the predicted trajectories, they may fail to capture critical events, such as collisions. 
To address this, we assume the agent moves at a constant speed between adjacent time steps and interpolate the low-frequency predicted trajectories to obtain high-frequency simulated motion trajectories, i.e., $T_{sim} > T_{pred}$.

\subsubsection{\textbf{Static Map}}
The static map module uses the initial map information from the dataset and assumes it remains unchanged during the simulation process. 
Specifically, the map information defines the road topology, consisting of lane markings that are represented as a series of polylines.
These polylines include solid, double solid, and broken lines, which encode different semantic properties of the lane markings.

\subsubsection{\textbf{Reward Function}}
The reward function calculates the immediate reward $r_t^{sim}$ for planned actions based on simulated future trajectories of the ego vehicle and other agents, combined with map information.
Specifically, the reward for a given simulation step $r_t^{sim}$ is composed of four components: collision reward $r_t^{col}$, lane-keeping reward $r_t^{lk}$, progress reward $r_t^{prog}$, and distance reward $r_t^{dist}$:
\begin{equation}
\begin{gathered}
r_t^{sim}=r_t^{col}+r_t^{lk}+r_t^{prog}+r_t^{dist}\\
\end{gathered}
\end{equation}
The collision reward $r_t^{col}$ is triggered by bounding-box overlap between the ego vehicle and other agents, and is set to -30 for vehicles, -50 for pedestrians, and -10 for traffic cones.
The lane-keeping reward $r_t^{lk}$ penalizes the ego for violating drivable boundaries, with penalties of -30 for crossing double solid lines and -10 for crossing single solid lines.
The progress reward $r_t^{{prog}} \in [0, 1]$ is a normalized value quantifying the percentage of the lateral path completed by the ego and is only given at the final step.
In addition, given the challenge of designing a comprehensive rule-based reward function solely from raw dataset annotations, we introduce a distance reward $r_t^{dist}$ as the negative $L_2$ distance between $P_{ego}^{sim}$ and the ground-truth future.
This implicitly models and encourages driving correctness, such as deceleration at stop signs.
The total reward $R^{sim}$ is defined as the weighted sum of the rewards $r_t^{sim}$ at each simulation step:
\begin{equation}
\begin{gathered}
R^{sim}=\sum_{t=1}^{T_{sim}} \gamma^{t-1} r_t^{sim}\\
\end{gathered}
\end{equation}
where $\gamma$ balances short-term and long-term simulation rewards, accounting for the impact of future action uncertainty during closed-loop testing.

\subsection{End-to-End Autonomous Driving Model}

The model structure of PerlAD is shown in Fig. \ref{overview}(c). 
As a modular E2E system, PerlAD integrates a perception encoder and unified transformer blocks for extracting structured representations from surround-view camera input $\mathcal{X}$.
Based on the structured representations, the Decoupled Planner (\text{DeP}) outputs decoupled actions $\{ a_{lat}, a_{lon} \}$, and the Prediction World Model (\text{PWM}) generates reactive agent trajectories.

\subsubsection{\textbf{Sparse Perception}}
To efficiently extract structured environmental representations from raw visual observations, we adopt the perception encoder from SparseDrive~\cite{sun2024sparsedrive}. 
This is centered around two learnable queries: agent queries $Q_a \in \mathbb{R}^{N_a \times D}$, designed to extract features of surrounding agents, and map queries $Q_m \in \mathbb{R}^{N_m \times D}$, to capture map elements. 
Here, $N_a$ and $N_m$ denote the number of queries, and $D$ is the feature dimension.
These queries $Q$ aggregate information from the high-dimensional inputs $\mathcal{X}$ and shift their corresponding anchors $\beta$ accordingly.
Specifically, the agent query uses box anchor $\beta_a \in \mathbb{R}^{N_a \times D_a}$ and map query uses polyline anchor $\beta_m \in \mathbb{R}^{N_m \times D_m}$, where $D_a$ and $D_m$ are anchor dimension.
The shifted anchors are used for downstream perception tasks, including object detection and mapping.

\subsubsection{\textbf{Unified Transformer Blocks}}
The output queries from sparse perception are then fed into Unified Transformer Blocks to model temporal-spatio interactions.
In addition to $Q_a$ and $Q_m$, we introduce decoupled ego planning queries $Q_{lat} \in \mathbb{R}^{1 \times D}$ and $Q_{lon} \in \mathbb{R}^{1 \times D}$ for lateral and longitudinal planning.
They are initialized from the front camera’s smallest feature map.
The corresponding ego planning anchor $\beta_e \in \mathbb{R}^{1 \times D_a}$ is represented in box format.
The complete agent-level query, $Q_{all} = \text{Concat}(Q_a, Q_{lat}, Q_{lon})$, is iteratively refined through $L$ attention modules. 
In each iteration $l \in [1,L]$, the query is updated sequentially for temporal-spatial interactions via:
\begin{equation}
\begin{aligned}
Q_{all}^l &\leftarrow \text{CrossAttn}(Q_{all}^{l-1}, Q_{all}^{hist}, Q_{all}^{hist}) \ \text{(Temporal)} \\
Q_{all}^l &\leftarrow \text{SelfAttn}(Q_{all}^l, Q_{all}^l, Q_{all}^l) \ \text{(Agent-Agent)}\\
Q_{all}^l &\leftarrow \text{CrossAttn}(Q_{all}^l, Q_{m}, Q_{m}) \ \text{(Agent-Map)}\\
\end{aligned}
\end{equation}
where \text{CrossAttn} and \text{SelfAttn} represent cross and self attentions, and $Q_{all}^{0} = Q_{all}$. 
The anchors $\beta$ serve as positional embeddings throughout the computation. 
After $L=3$ iterations, the updated agent-level queries are denoted as $\hat{Q}_{all} = \{ \hat{Q}_{a}, \hat{Q}_{lat},\hat{Q}_{lon} \}$, where the decoupled ego planning queries $\hat{Q}_{lat}$ and $\hat{Q}_{lon}$ are leveraged by $\text{DeP}$ to perform lateral path planning and longitudinal speed planning.
The agent queries $\hat{Q}_a$ are given to $\text{PWM}$ for reactive predictions.

\begin{figure}[htp]
\centering
\includegraphics[width=0.44\textwidth]{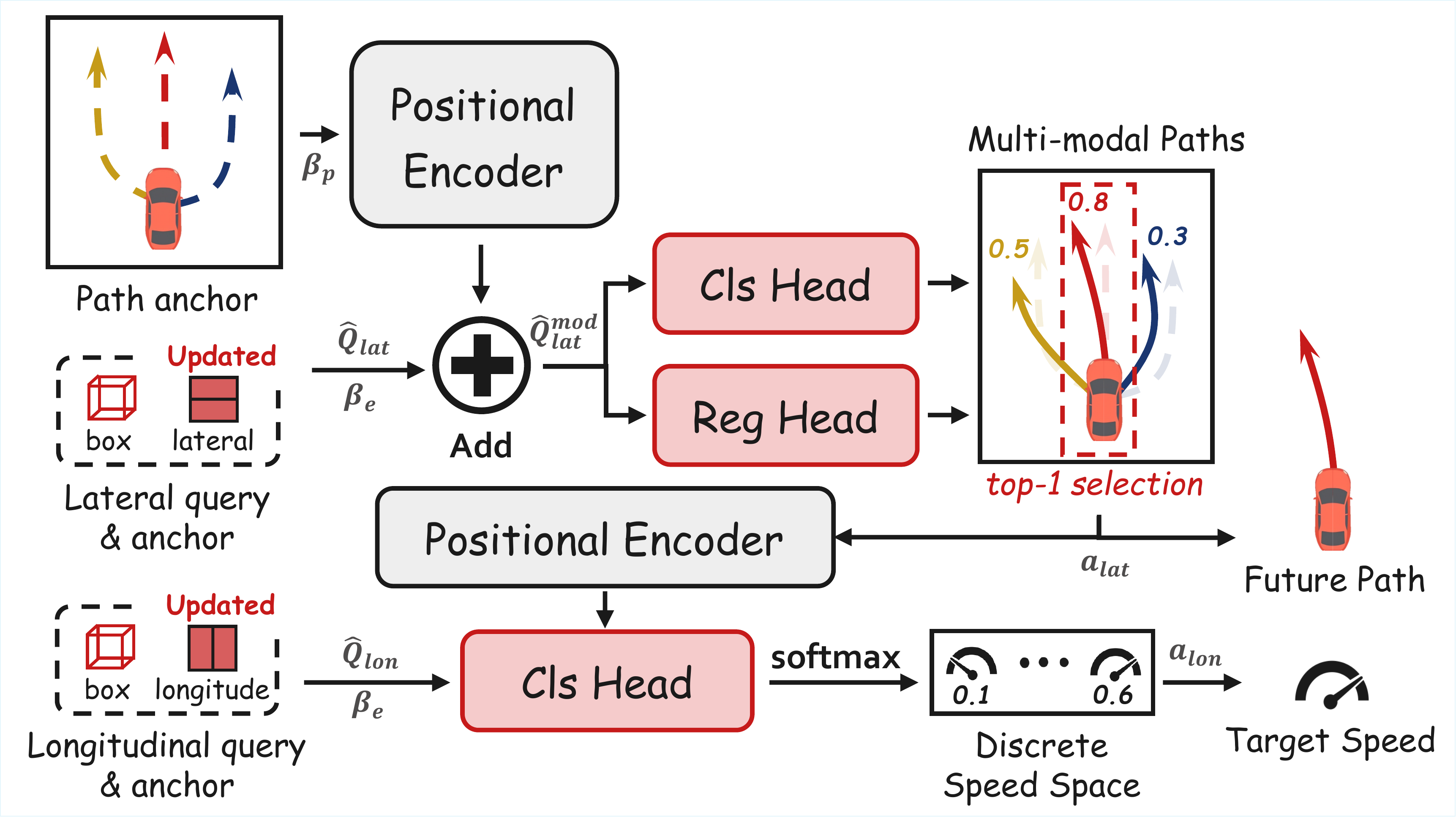}
\vspace{-3mm}
\caption{PerlAD adopts a hierarchical decoupled planning scheme: the lateral planner outputs multi-modal future paths, while the longitudinal planner generates target speeds conditioned on the selected path.}
\label{planner fig}
\vspace{-6mm}
\end{figure}

\subsubsection{\textbf{Decoupled Planner (\text{DeP})}}
\label{planner}

Most E2E methods adopt a coupled planning output, similar to agent trajectory prediction~\cite{sun2024sparsedrive}.
Despite being easy to deploy, the coupled planning output often leads to suboptimal lateral control~\cite{jaeger2023hidden}.
As shown in Fig. \ref{planner fig}, PerlAD adopts a hierarchical, decoupled planning approach, where the lateral path $a_{lat}$ and longitudinal speed $a_{lon}$ are output sequentially.
Specifically, $a_{lat}$ determines the ego's geometric path and driving intention (e.g., lane changing or lane keeping), while $a_{lon}$ focuses on optimizing speed control to handle dynamic interactions and ensure safety.

\textbf{Lateral Planning.}
The lateral action $a_{lat}=\{ w_{ego, i} \}_{i=1}^{N_{lat}}$ determines the ego vehicle's future spatial path.
To capture diverse driving intentions, this branch adopts a multi-modal path planning approach.
Specifically, the lateral planning query $\hat{Q}_{lat}$ is augmented by incorporating both the ego planning anchor $\beta_e$ and pre-clustered path anchors $\beta_p$:
\begin{equation}
\begin{aligned}
\hat{Q}_{lat}^{mod} = \hat{Q}_{lat}+\text{PE}(\beta_e)+\text{PE}(\beta_p)
\end{aligned}
\end{equation}
This results in an expanded multi-modal path query $\hat{Q}_{lat}^{mod} \in \mathbb{R}^{1 \times K_{path} \times D}$, where $K_{path}$ is the number of path modalities and $\text{PE}$ denotes positional encoding.
A regression head and a classification head are then applied to $\hat{Q}_{lat}^{mod}$ to output multiple possible paths and their associated probability scores jointly. 
The final lateral action $a_{lat}$ is selected as the path with the highest probability score. 

\textbf{Longitudinal Planning.}
The longitudinal action $a_{lon}$ is a scalar that determines the ego’s target speed.
The target speed is defined over a discrete action space with cardinality $K_{speed}$.
Specifically, the longitudinal planning query $\hat{Q}_{lon}$ is augmented by incorporating both the ego planning anchor $\beta_e$ and previously selected lateral path $a_{lat}$ via positional encoding.
This augmented query is then input to a classification head, which produces a softmax probability distribution over the discrete action space. 
The action with the highest probability is selected as the final longitudinal action $a_{lon}$.

\begin{figure}[htp]
\vspace{-3mm}
\centering
\includegraphics[width=0.44\textwidth]{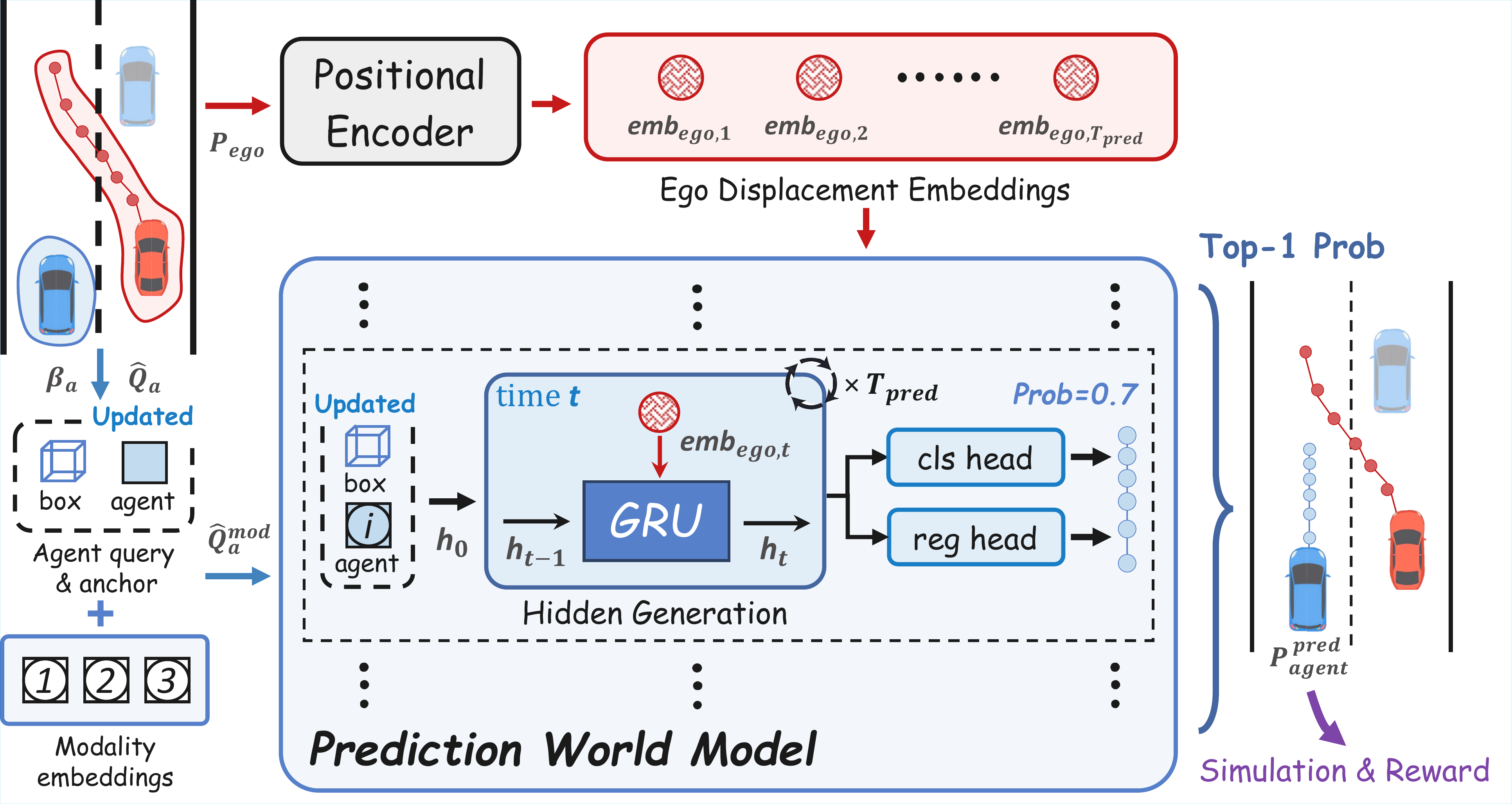}
\vspace{-3mm}
\caption{
The Prediction World Model autoregressively generates multi-modal trajectory predictions for surrounding agents, explicitly conditioned on the ego's future trajectory.
The highest-probability modality is selected for subsequent simulation and reward computation.}
\label{synthetic reward}
\vspace{-3mm}
\end{figure}

\subsubsection{\textbf{Prediction World Model (\text{PWM})}}
\label{world model}
Traditional E2E models often neglect the ego vehicle's behavior during motion prediction~\cite{jiang2023vad, jia2025drivetransformer}.
This hinders the simulation of reactive scenarios critical for closed-loop driving.
PerlAD introduces $\text{PWM}$, which explicitly conditions agents' prediction processes on the ego's future trajectory, thereby enabling reactive prediction.
Fig. \ref{synthetic reward} illustrates its network design and workflow.

\textbf{Ego-Conditional Prediction.}
PWM employs a generative network to predict agent trajectories~\cite{zheng2024genad}, which autoregressively generates agents' hidden states with a gated recurrent unit ($\text{GRU}$).
To account for multi-modality, agent queries are first augmented with modality embeddings, resulting in expanded agent queries $\hat{Q}_a^{mod} \in \mathbb{R}^{N_a \times K_{pred} \ \times D}$, where $K_{pred}$ is the number of prediction modalities.
The initial hidden state $h_0$ for the $\text{GRU}$ is derived from the combination of $\hat{Q}_a^{mod}$ and their anchors' position embeddings:
\begin{equation}
\begin{aligned}
h_{0} = \hat{Q}_a^{mod}+\text{PE}(\beta_a)
\end{aligned}
\end{equation}
At each time step $t$, the hidden state is updated by incorporating the ego vehicle's displacement embedding.
This embedding captures the conditional information from the ego trajectory $P_{ego}=\{ w_{ego,t} \}_{t=1}^{T}$:
\begin{equation}
\begin{aligned}
emb_{ego, t} &= \text{PE}(w_{ego, t} - w_{ego, t-1})\\
h_t &= \text{GRU}(h_{t-1}, emb_{ego,t})
\end{aligned}
\end{equation}
Note that during PWM’s supervised training, the ego trajectory $P_{ego}$ is taken from the ground truth.
However, when the PWM is used to provide reactive simulation for RL training, $P_{ego}$ is generated by giving \text{DeP}'s output actions to the ego simulation.
Finally, based on the multi-modal hidden states, $\text{PWM}$ predicts multiple trajectories and their associated probabilities with regression and classification heads, selecting the trajectory with the highest probability as its output $P_{agent}^{pred}$.

\textbf{Simulating Reactive Scenarios.}
During RL training, the policy interacts with the pseudo-simulation to collect rewards. 
To ensure consistency between the interaction behaviors of simulated agents and those expected in closed-loop testing scenarios, the future trajectories of surrounding agents in the pseudo-simulation are derived from the reactive predictions generated by \text{PWM}.
Specifically, given a low-frequency Top-1 predicted trajectory $P_{agent}^{pred}$, the pseudo-simulation transforms it into a high-frequency simulated trajectory $P_{agent}^{sim}$.

\subsection{Training Strategies}
To facilitate efficient and stable model convergence, PerlAD adopts a two-stage training approach.
In the first stage, only the sparse perception is trained using detection and mapping losses, allowing the model to learn structured scene representations from raw sensor inputs:
\begin{equation}
\mathcal{L}_{stage1} = \mathcal{L}_{det} + \mathcal{L}_{map}    
\end{equation}
In the second stage, the pretrained sparse perception is frozen. 
The unified transformer blocks, $\text{DeP}$, and $\text{PWM}$ are jointly optimized with prediction and planning losses:
\begin{equation}
\mathcal{L}_{stage2} = \mathcal{L}_{pred} + \mathcal{L}_{plan}
\end{equation}
The prediction loss $\mathcal{L}_{pred}$ supervises the PWM to generate accurate predictions.
The planning loss $\mathcal{L}_{plan}$ is further decoupled into lateral $\mathcal{L}_{lat}$ and longitudinal $\mathcal{L}_{lon}$ components.

\subsubsection{\textbf{IL-based Lateral Planning Training}}
To generate smooth and continuous paths, the lateral planning component is trained primarily with dense supervision provided by IL, rather than relying on RL to explore the high-dimensional continuous coordinate space, which would significantly increase the optimization difficulty.
The lateral loss is formulated as:
\begin{equation}
\mathcal{L}_{lat} = \mathcal{L}_{lat}^{reg} + \mathcal{L}_{lat}^{cls}
\end{equation}
where ${L}_{lat}^{reg}$ and ${L}_{lat}^{cls}$ correspond to the regression and classification objectives for multi-modal path output, respectively.

\subsubsection{\textbf{RL-based Longitudinal Planning Training}}
Conditioned on the lateral path, the longitudinal branch controls the ego’s interactions with the environment via speed planning. 
Since IL lacks explicit feedback and exploration over interactive outcomes, we employ the $\text{REINFORCE}$ method~\cite{sutton2018reinforcement} with group-standardized advantage estimation~\cite{guo2025deepseek} to conduct RL training. 
Specifically, for a given longitudinal query $\hat{Q}_{lon}$ and lateral path $a_{lat}$, we sample $G$ target speeds $\{ a_{lon, i} \}_{i=1}^{G}$ and execute them in the pseudo-simulaton to obtain a set of corresponding rewards $\{ R_{i}^{sim} \}_{i=1}^{G}$.
The policy $\pi$ is then updated according to $\mathcal{L}_{lon}$:
\begin{equation}
\mathcal{L}_{lon} = - \frac{1}{G}\left[\sum_{i=1}^{G} \log \pi(a_{lon, i}|\hat{Q}_{lon},a_{lat}) \cdot A_i\right] - \mathcal{L}_{ent}
\end{equation}
where $A_i=(R_{i}^{sim}-\text{mean}(\{ R_{j}^{sim} \}_{j=1}^{G})) /\ \text{std}(\{ R_{j}^{sim} \}_{j=1}^{G})$ is the estimated advantage, $\mathcal{L}_{ent}$ is the entropy loss.

\subsubsection{\textbf{Lateral-Longitudinal Alignment}}
We use a curriculum strategy to align the decoupled planning branches. 
In the initial phase of stage 2, before lateral planning converges, the path input for longitudinal planning is derived from the ground truth.
As training progresses into the final third, when the lateral path prediction has achieved sufficient accuracy, the input path switches to the predicted path, aligning lateral and longitudinal planning.
Furthermore, to better synchronize two planning branches, we modify the lateral path classification loss $\mathcal{L}_{lat}^{cls}$ by augmenting the selection criteria.
Specifically, instead of selecting the best path solely based on the distance to the ground truth, we additionally incorporate clipped rewards.
This ensures the model selects paths that are both geometrically accurate and lead to safer, more efficient maneuvers.

\subsubsection{\textbf{Reactive Training Simulation}}
To avoid using incorrect reward signals from low-quality predictions during the early stages of training, we simulate the behavior of surrounding agents using ground truth trajectories at the beginning.
As training progresses into the final third, when the PWM predictions become stable and sufficiently accurate, we progressively replace ground truth trajectories with PWM's reactive predictions, thereby providing more reliable reward signals.

\section{Experiments}

\begin{table*}[htbp]
\centering
\small
\caption{\textbf{Closed-loop and Multi-ability Results of E2E-AD Methods on Bench2Drive Leaderboard.}}
\vspace{-2mm} 
\label{tab: b2d}
\resizebox{0.88\linewidth}{!}
{
\begin{threeparttable}
\begin{tabular}{lc|>{\columncolor[gray]{0.9}}c>{\columncolor[gray]{0.9}}ccc|ccccc|>{\columncolor[gray]{0.9}}c}
\toprule
\multirow{2}{*}{\textbf{Methods}} & \multirow{2}{*}{\textbf{Reference}} & \multicolumn{4}{c|}{\textbf{Closed-loop Metric} $\uparrow$ } & \multicolumn{6}{c}{\textbf{Multi-ability (\%)} $\uparrow$} \\ 
\cmidrule{3-12} 
& & DS & SR (\%) & Efficiency & Comfortness & Merging & Overtaking & Emergency Brake & Give Way & Traffic Sign & \textbf{Mean} \\ 
\midrule

UniAD~\cite{hu2023planning} & CVPR 23 & 45.81 & 16.36 & 129.21 & 43.58 & 12.16 & 20.00 & 23.64 & 10.00 & 13.89 & 15.89 \\
VAD~\cite{jiang2023vad} & ICCV 23 & 42.35 & 15.00 & 157.94 & 46.01 & 7.14 & 20.00 & 16.36 & 20.00 & 20.22 & 16.75 \\
SparseDrive~\cite{sun2024sparsedrive} & ICRA 25 & 44.54 & 16.71 & 170.21 & 48.63 & 12.18 & 23.19 & 17.91 & 20.00 & 20.98 & 17.45  \\
DriveTrans.-L.~\cite{jia2025drivetransformer} & ICLR 25 & 63.46 & 35.01 & 100.64 & 20.78 & 17.57 & 35.00 & 48.36 & 40.00 & 52.10 & 38.60 \\
DiffAD~\cite{wang2025diffad} & arXiv 25 & 67.92 & 38.64 & - & - & 30.00 & 35.55 & 46.66 & 40.00 & 46.32 & 38.79 \\
WOTE~\cite{li2025wote} & ICCV 25 & 61.71 & 31.36 & - & - & - & - & - & - & - & - \\
Hydra-Next~\cite{li2025hydra} & ICCV 25 & \underline{73.86} & 50.00 & 197.76 & 20.68 & 40.00 & \underline{64.44} & \underline{61.67} & 50.00 & 50.00 & 53.22 \\
ReasonPlan~\cite{liu2025reasonplan} & CoRL 25 & 64.01 & 34.55 & 180.64 & 25.63 & 37.50 & 26.67 & 33.30 & 40.00 & 45.79 & 36.66 \\
Raw2Drive~\cite{yang2025raw2drive} & NeurIPS 25 & 71.36 & \underline{50.24} & 214.17 & 22.42 & \textbf{43.35} & 51.11 & 60.00 & 50.00 & \textbf{62.26} & \underline{53.34} \\
TakeAD~\cite{liu2026takead} & RA-L 26 & 71.39 & 40.83 & 193.30 & 22.89 & 30.77 & 35.56 & 56.67 & 50.00 & 42.02 & 43.00 \\
\midrule
PerlAD (\textbf{Ours}) & - & \textbf{78.70} & \textbf{57.27} & 181.16 & 22.37 & \underline{40.00} & \textbf{75.56} & \textbf{68.33} & \textbf{50.00} & \underline{52.11} & \textbf{57.20} \\ 
\bottomrule
\end{tabular}
\begin{tablenotes}
     \item[] Results are taken from the original papers. ``–'' indicates metrics that are not reported.
\end{tablenotes}
\vspace{-2mm}
\end{threeparttable}
}
\vspace{-4mm}
\end{table*}

\subsection{Experiment Setup}

\subsubsection{\textbf{Dataset and Benchmarks}}
We evaluate our model on the challenging Bench2Drive (B2D)~\cite{jia2024bench} benchmark.
The offline training dataset is B2D-Base, which consists of 1,000 video clips (approximately 230K frames) collected by the privileged expert in Think2Drive~\cite{li2024think2drive}. 
The corresponding evaluation dataset contains 12,806 frames.
For closed-loop evaluation, we follow the official B2D protocol and test on the 220 routes.
We conduct additional evaluations on Driving in Occlusion Simulation (DOS)~\cite{shao2023reasonnet} to assess the model's performance in safety-critical scenarios.
During closed-loop testing, the lateral action $a_{lat}$ is converted to steering commands, and the longitudinal action $a_{lon}$ to throttle and brake signals, via PID controllers identical to those in the pseudo-simulation. 
Both B2D and DOS benchmarks focus on urban driving in a low-speed regime, making our decoupled action design well-suited.

\subsubsection{\textbf{Metrics}}
We adopt the official evaluation metrics provided by B2D.
Key metrics include the Driving Score (DS), which measures overall performance by considering both route completion and traffic violations, and the Success Rate (SR), which reflects the percentage of routes that achieve the maximum DS.
Efficiency and Comfortness quantify the ego’s relative speed compared to surrounding agents and the smoothness of its motion, respectively.
More details about the metrics can be found in the original B2D paper~\cite{jia2024bench}.

\subsubsection{\textbf{Implementation Details}}
We adopt ResNet-50 as the backbone to extract visual features from $N_{cam}=6$ surround-view cameras.
Agent and map queries are set to $N_a=900$ and $N_m=100$. 
A sinusoidal positional encoder maps coordinates into high-dimensional space. 
All regression and classification heads are implemented as two-layer Multi-Layer Perceptrons (MLPs) with feature dimension $D=256$.
Navigation information (target points and one-hot driving commands) is embedded into path queries via two-layer MLPs.
We use $K_{pred}=6$ prediction modalities.
Path modalities $K_{path}=8$, each represented by $N_{lat}=6$ waypoints sampled every 2m. 
For longitudinal planning, the maximum speed is set to 12 m/s based on the speed distribution of training datasets, which is uniformly discretized into $K_{speed}=13$ levels. 
Agent prediction operates at 2Hz over 3s, while simulation runs at 10Hz over 2s, which makes $T_{pred}=6$ and $T_{sim}=20$. 
Training is conducted on 8 NVIDIA H20 GPUs with batch size 256 for 12 epochs in stage 1 at learning rate $4e^{-4}$ and 18 epochs in stage 2 at $2e^{-4}$, using AdamW optimizer with a weight decay of 0.01.
During RL training, $G=32$ speed actions are sampled per sample with discount factor $\gamma=0.9$.

\subsection{Results and Analysis}

\subsubsection{\textbf{Quantitative Comparison}}
We conduct comprehensive experiments to evaluate PerlAD’s driving performance.

\textbf{SoTA performance in complex interactive scenarios.}
Table~\ref{tab: b2d} summarizes the closed-loop results across all 220 routes of the B2D benchmark.
PerlAD achieves SoTA performance in the closed-loop metrics, surpassing existing IL-based methods. 
Compared with our IL baseline SparseDrive~\cite{sun2024sparsedrive}, PerlAD delivers a 76.7\% improvement in DS (+34.16) and a 40.56\% boost in SR.
This substantial gain is attributed to our reactive pseudo-simulation-based RL training, which effectively optimizes policy behavior towards real closed-loop driving objectives. 
Against the only published RL method Raw2Drive~\cite{yang2025raw2drive} that relies on costly online interactions and expert distillation, PerlAD still attains a 10.29\% higher DS (+7.34) without requiring online exploration. 
The multi-ability metrics reports the SR across specific categories of driving scenarios.
PerlAD achieves leading SR in Merging, Overtaking, and Emergency Brake scenarios, yielding the best mean SR and demonstrating its ability to capture interactive behaviors for reliable performance in complex environments.


\begin{table}[h]
\renewcommand{\arraystretch}{1.0}
\vspace{-4mm}
\caption{\textbf{Closed-loop Results on DOS.}
We report the Driving Score across four occlusion scenarios: Parked Cars (DOS\_01), Sudden Brake (DOS\_02), Left Turn (DOS\_03), and Red Light Infraction (DOS\_04).}
\vspace{-4mm}
\label{tab: dos}
\setlength\tabcolsep{2pt}
\begin{center}
\resizebox{0.76\linewidth}{!}{\begin{tabular}{l|c|c|c|c|>{\columncolor[gray]{0.9}}c}
\toprule
\multirow{2}{*}{\textbf{Methods}} & \multicolumn{5}{c}{\textbf{Driving Score (DS) in DOS $\uparrow$}} \\
\cmidrule{2-6}
& DOS\_01 & DOS\_02 & DOS\_03 & DOS\_04 & \textbf{Average} \\
\cmidrule{1-6}
UniAD~\cite{hu2023planning} & 66.00 & 69.32 & 73.18 & 75.87 & 71.09 \\
VAD~\cite{jiang2023vad} & 58.46 & 57.84 & 61.16 & 65.88 & 60.84 \\
LMDrive~\cite{shao2024lmdrive} & 64.00 & 62.00 & 73.65 & 58.54 & 64.53 \\
ReasonPlan~\cite{liu2025reasonplan} & \underline{67.77} & \textbf{91.65} & \underline{80.57} & \underline{77.85} & \underline{78.02} \\
\midrule
PerlAD (\textbf{Ours}) & \textbf{87.71} & \underline{87.00} & \textbf{87.40} & \textbf{85.21} & \textbf{86.83} \\
\hline 
\end{tabular}}
\vspace{-4mm}
\end{center}
\end{table}

\textbf{Additional evaluation in safety-critical occlusion scenarios.}
To assess PerlAD's performance under high-risk conditions, we extend evaluations to the Driving in Occlusion Simulation (DOS) benchmark~\cite{shao2023reasonnet}, which characterizes four categories of occlusion scenarios and serves as a rigorous testbed for safety-critical planning.
As shown in Table \ref{tab: dos}, PerlAD achieves the highest average DS, indicating RL objective mitigates the inherent limitations of IL in hazardous situations.
By explicitly optimizing for collision-related rewards within the pseudo-simulation, PerlAD develops a strong safety-aware planning ability that is crucial for occlusion scenarios.

\begin{table}[htbp]
\centering
\vspace{-2mm}
\caption{\textbf{Improved prediction accuracy from the Prediction World Model}. 
ADE / FDE = Average / Final Displacement Error.
} 
\vspace{-2mm}
\label{PWM accuracy}
\resizebox{0.83\linewidth}{!}{
\begin{tabular}{l|C{0.7cm}|C{0.7cm}|C{0.7cm}|C{0.7cm}} 
\toprule
\multirow{2}{*}{\textbf{Methods}} & \multicolumn{2}{c|}{\textbf{Best-of-K (m) $\downarrow$}} & \multicolumn{2}{c}{\textbf{Top-1 (m)} $\downarrow$} \\
\cmidrule{2-5}
& $\text{ADE}$ & $\text{FDE}$ & $\text{ADE}$ & $\text{FDE}$ \\ 
\cmidrule{1-5}
Vanilla, w/o. action & 0.69 & 0.97 & 1.44 & 2.56 \\
\cmidrule{1-5}
PWM, w. predicted action & \underline{0.64} & \underline{0.87} & \underline{1.31} & \underline{2.31} \\
PWM, w. ground truth action & \textbf{0.62} & \textbf{0.85} & \textbf{1.29} & \textbf{2.24} \\
\bottomrule
\end{tabular}}
\vspace{-3mm}
\end{table}

\textbf{Improved prediction accuracy and reactivity via $\text{PWM}$.}
We analyze the advantages of PWM from its improved prediction accuracy and reactivity on B2D evaluation datasets.
In Table \ref{PWM accuracy}, we report prediction accuracy for vehicle agents using Best-of-K results for multi-modal prediction~\cite{sun2024sparsedrive}, and Top-1 results used for simulation.
Compared with vanilla single-shot non-reactive prediction, \text{PWM} with predicted actions achieves higher prediction accuracy. 
Using ground-truth actions that align with the prediction target further reduces the discrepancy, suggesting that \text{PWM} captures correlations between ego actions and agent behaviors.

\begin{table}[htbp]
\centering
\vspace{-3mm}
\caption{\textbf{Improved reactivity from the Prediction World Model}. 
} 
\vspace{-2mm}
\label{PWM reactivity}
\resizebox{0.86\linewidth}{!}{
\begin{tabular}{l|c|c|c} 
\toprule
\multirow{3}{*}{} & \multicolumn{3}{c}{\textbf{Counterfactual Collision Numbers $\downarrow$}}\\
\cmidrule{2-4}
\textbf{Methods} & High-speed $\rightarrow$ & Stationary $\rightarrow$ & Total \\
& Sudden Brake & Abrupt Lane-change & \\ 
\cmidrule{1-4}
Logged Replay & 144 / 160 & 22 / 40 & 166 / 200 \\
Vanilla, w/o. action & 124 / 160 & 23 / 40 & 147 / 200\\
PWM & 57 / 160 & 13 / 40 & 70 / 200 \\
\bottomrule
\end{tabular}}
\vspace{-3mm}
\end{table}

To validate PWM's ability to generate reactive predictions, inspired by prior works on simulation agents~\cite{zhang2022trajgen}, we introduce the \textit{counterfactual evaluation} based on B2D evaluation datasets, where we preserve the ego’s ground-truth future path while modifying only its target speed, and quantify reactivity by measuring whether predicted agent trajectories collide with the ego vehicle within a 2s simulation.
The evaluation comprises 200 valid cases generated by first constructing two types of counterfactual scenarios and then filtering out scenarios in which agent responses show low relevance to ego actions.
This yields 160 sudden brake cases where the ego decelerates from speeds above 7.5 m/s to a modified target speed of 0 m/s on straight roads,
and 40 abrupt lane-change cases starting from a stationary state of 0 m/s to a modified target speed of 9 m/s.
Original target speeds remain high and zero, respectively, ensuring genuine counterfactual evaluations.

As shown in Table \ref{PWM reactivity}, \text{PWM} achieves significantly fewer collisions, validating that it captures interaction-dependent dynamics beyond static trajectory prediction, leading to more reactive simulated scenarios.
Although \text{PWM} supports counterfactual prediction under unlogged ego actions, it does not explicitly model extreme adversarial behaviors, highlighting a potential extension to the current framework.

\subsubsection{\textbf{Ablation Study}}
Following the official recommendation~\cite{jia2025drivetransformer}, we conduct our ablation study on \textit{Dev10}, a subset of the 220 routes comprising 10 challenging and representative routes.
For clarity, we report the main metrics of DS and SR. 
We also introduce the Collision Rate (CR) to emphasize safety performance, calculated as the number of collisions per hundred meters traveled by the ego.

\textbf{Effects of training strategies.}
In this work, we propose three specialized training strategies to obtain a high-performance E2E policy: 
RL training for longitudinal speed control (RL-lon), lateral-longitudinal alignment (LLA), and reactive training simulation (RTS). 
Specifically, RL-lon replaces IL for longitudinal planning to acquire interactive speed outputs. 
The LLA links the decoupled planning branches by conditioning the longitudinal planning on the predicted lateral path, and by adjusting the lateral path modality probability using RL rewards. 
RTS provides reactive simulations via $\text{PWM}$, thereby ensuring that the RL training reward is consistent with the dynamic closed-loop interactions.

\begin{table}[htbp]
\centering
\vspace{-2mm}
\caption{\textbf{Ablation on proposed training strategies.} LLA and RTS denote lateral-longitudinal alignment and reactive training simulation, respectively.} 
\vspace{-2mm}
\label{ablation training}
\resizebox{0.85\linewidth}{!}{
\begin{tabular}{cccc|ccc} 
\toprule
\multicolumn{4}{c|}{\textbf{Training Strategies}} & \multicolumn{3}{c}{\textbf{Closed-loop Metric}}  \\ 
\cmidrule{1-7}
IL-lon & RL-lon & LLA & RTS & DS $\uparrow$ & SR (\%) $\uparrow$ & CR $\downarrow$ \\ 
\cmidrule{1-7}
\checkmark & $\times$ & $\times$ & $\times$ & 32.81 & 0 & 0.74 \\
$\times$ & \checkmark & $\times$ & $\times$ & 65.01 & 20 & 0.39 \\
$\times$ & \checkmark & \checkmark & $\times$ & 70.28 & 40 & 0.20 \\
$\times$ & \checkmark & \checkmark & \checkmark & 74.00 & 40 & 0.09 \\
\bottomrule
\end{tabular}}
\vspace{-3mm}
\end{table}

We also introduce the IL version of longitudinal planning (IL-lon), which uses two-hot label classification to imitate target speed actions. 
As shown in Table \ref{ablation training}, IL-lon achieves markedly low DS compared to RL-lon.
This is primarily due to a high rate of collision penalties or getting stuck at the starting point, which suggests that IL alone is insufficient for achieving interactive speed control, highlighting the necessity of RL training.
Further introducing LLA upon RL-lon significantly improves the joint performance of the decoupled planning in closed-loop testing. 
Finally, the incremental performance gain achieved by introducing RTS demonstrates its critical role by providing reliable reactive simulations.

\textbf{Effects of reward function design.}
The reward function contains four types of components: the collision reward $r^{col}$, the lane-keeping reward $r^{lk}$, the progress reward $r^{prog}$, and the distance reward $r^{dist}$.

\begin{table}[htbp]
\centering
\vspace{-2mm}
\caption{\textbf{Ablation on reward function design.}
Total reward includes collision $r^{col}$, lane-keeping $r^{lk}$, progress $r^{prog}$, and distance $r^{dist}$.} 
\vspace{-2mm}
\label{ablation reward}
\resizebox{0.87\linewidth}{!}{
\begin{tabular}{c|cccc|ccc} 
\toprule
\multirow{2}{*}{\textbf{ID}} & \multicolumn{4}{c|}{\textbf{Reward Terms}} & \multicolumn{3}{c}{\textbf{Closed-loop Metric}}  \\ 
\cmidrule{2-8}
& $r^{col}$ & $r^{lk}$ & $r^{prog}$ & $r^{dist}$ & DS $\uparrow$ & SR (\%) $\uparrow$ & CR $\downarrow$ \\ 
\cmidrule{1-8}
1 & $\times$ & $\times$ & $\times$ & \checkmark & 53.58 & 20 & 0.83 \\
2 & \checkmark & $\times$ & $\times$ & \checkmark & 63.54 & 20 & 0.41 \\
3 & \checkmark & \checkmark & $\times$ & \checkmark & 67.26 & 40 & 0.21\\
4 & \checkmark & \checkmark & \checkmark & $\times$ & 65.66 & 30 & 0.31 \\
5 & \checkmark & \checkmark & \checkmark & \checkmark & 74.00 & 40 & 0.09 \\
\bottomrule
\end{tabular}}
\vspace{-3mm}
\end{table}

Table~\ref{ablation reward} demonstrates the necessity and incremental benefits of each reward component.
The baseline configuration (ID~1), relying solely on the distance reward $r^{dist}$, yields a low DS due to its neglect of critical driving constraints. 
The introduction of collision rewards $r^{col}$ (ID~2) and lane-keeping rewards $r^{lk}$ (ID~3) provides consistent performance gains by explicitly penalizing collisions and improving path adherence. 
Incorporating the progress reward (ID~5) achieves the best performance, validating the efficacy of jointly optimizing these complementary objectives. 
Notably, although a feasible policy can still be learned without $r^{dist}$ (ID~4), removing this auxiliary signal hinders the model’s ability to capture implicit behaviors not fully encoded by rule-based rewards (e.g., obeying traffic signs), resulting in lower performance compared to the full configuration.

\begin{figure}[htp]
\vspace{-3mm}
\centering
\includegraphics[width=0.42\textwidth]{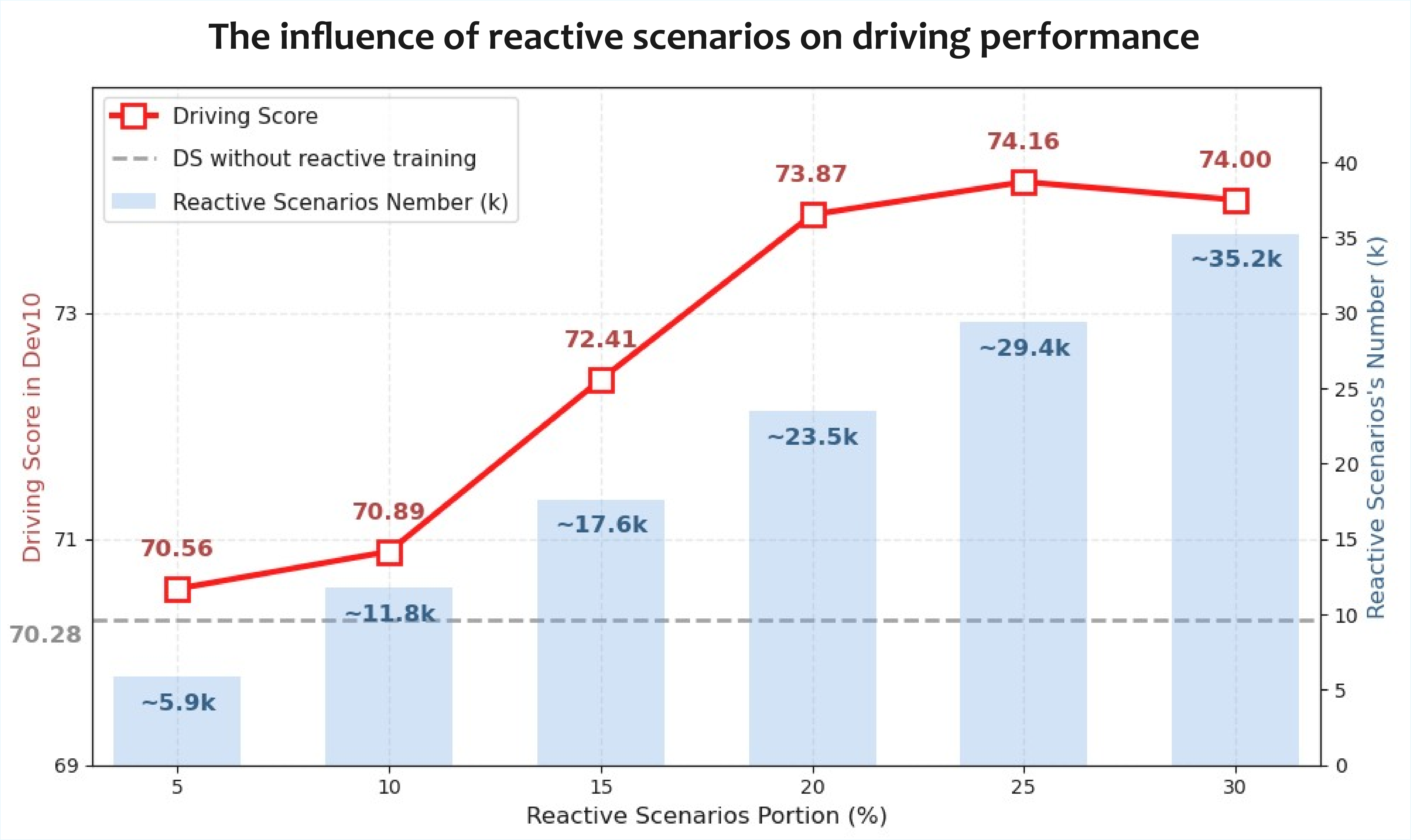}
\vspace{-3mm}
\caption{Ablation on the influence of reactive scenarios portions and cumulative reactive scenarios number.
Driving Score on Dev10 is reported.}
\label{ablation reactive training}
\vspace{-3mm}
\end{figure}

\textbf{Effects of reactive scenarios.}
During RL training, the agent’s behavior is gradually shifted from static logged trajectories to the $\text{PWM}$'s reactive predictions.
As shown in Fig. \ref{ablation reactive training}, the x-axis represents the proportion of training samples that use $\text{PWM}$'s predictions to simulate driving scenarios and calculate safety rewards. 
We also report the cumulative number of these reactive samples.
The results show that policy performance improves notably as the proportion of reactive scenarios increases, indicating the closed-loop performance gains from the reactive training process.
Note that although the observation inputs for the pseudo-simulation are drawn from the offline datasets, explicitly modeling the reactive behaviors of surrounding agents enables the policy to be trained under interactions that are better consistent with closed-loop execution, including interactions that go beyond the static replay.

As the proportion of reactive scenarios increases to around $30\%$, the performance gains gradually saturate, revealing the inherent limitations of offline logged observations due to their insufficient coverage of the online environment.
A promising direction for future improvement is to incorporate a latent world model operating in the sensory feature space, enabling more efficient extrapolation beyond the offline data distribution and further enhancing the RL agent's capability.

\section{Conclusion}
In this work, we present PerlAD, a novel Reinforcement Learning (RL) training framework for enhanced closed-loop End-to-end (E2E) autonomous driving. 
We overcome existing limitations through three core innovations: a data-driven pseudo-simulation for efficient rendering-free RL interaction, a Prediction World Model that generates reactive agent behaviors consistent with closed-loop scenarios, and a hierarchical decoupled planner with an alignment strategy for joint lateral-longitudinal optimization.
Comprehensive experiments demonstrate that PerlAD achieves state-of-the-art closed-loop performance on the challenging Bench2Drive benchmark, with strong safety validated on DOS.
Future work includes developing a latent world model for efficient extrapolation of sensory features beyond offline training data, and exploring reward modeling based on human preference data to build a more comprehensive reward function, abandoning the distance term. 
Additionally, although the decoupled planning paradigm shows promising experimental results in low-speed regimes, extending RL toward coupled planning optimization that is more generalizable to diverse driving conditions remains an important research direction.

\bibliographystyle{IEEEtran}
\bibliography{ref} 

\end{document}